%% file: main.tex
\documentclass[10pt,twocolumn,letterpaper]{article}

\usepackage{iccv}
\usepackage{times}
\usepackage{epsfig}
\usepackage{graphicx}
\usepackage{amsmath}
\usepackage{amssymb}

\makeatletter
\@namedef{ver@everyshi.sty}{}
\makeatother

\usepackage{booktabs}

\usepackage{multirow}
\usepackage{ulem}
\usepackage{amsmath}
\usepackage{bbm}
\usepackage{stmaryrd}

\usepackage[dvipsnames]{xcolor}
\usepackage{enumitem}
\usepackage{microtype}
\usepackage{nicematrix}

\newenvironment{packed_lefty_item}{
\begin{itemize}[leftmargin=*]
\vspace{-6pt}
  \setlength{\itemsep}{0pt}
  \setlength{\parskip}{0pt}
  \setlength{\parsep}{0pt}
  \setlength{\topsep}{-10pt}
  \setlength{\partopsep}{0pt}
}{\end{itemize}\vspace{-6pt}}

\usepackage[pagebackref,breaklinks,colorlinks]{hyperref}

\usepackage[capitalize]{cleveref}
\crefname{section}{Sec.}{Secs.}
\Crefname{section}{Section}{Sections}
\Crefname{table}{Table}{Tables}
\crefname{table}{Tab.}{Tabs.}

\iccvfinalcopy 

\def\iccvPaperID{4874} 

\ificcvfinal\fi

\begin{document}

\title{GLOW: Global Layout Aware Attacks on Object Detection}

\author{Jun Bao\textsuperscript{*}\\
{\tt\small baoj@zju.edu.cn}
\and
Buyu Liu\textsuperscript{*}\\
{\tt\small buyu@nec-labs.com}
\and
Jianping Fan\\
{\tt\small jfan1@Lenovo.com}\\
\and
Xi Peng\\ 
{\tt\small pengx.gm@gmail.com}\\
\and
Kui Ren\\
{\tt\small kuiren@zju.edu.cn}\\
\and
Jun Yu\\
{\tt\small yujun@hdu.edu.cn}
}

\maketitle
\ificcvfinal\thispagestyle{empty}\fi

\begin{abstract}
Adversarial attacks aim to perturb images such that a predictor outputs incorrect results. Due to the limited research in structured attacks, imposing consistency checks on natural multi-object scenes is a promising yet practical defense against conventional adversarial attacks. More desired attacks, to this end, should be able to fool defenses with such consistency checks. Therefore, we present the first approach GLOW that copes with various attack requests by generating global layout-aware adversarial attacks, in which both categorical and geometric layout constraints are explicitly established. Specifically, we focus on object detection task and given a victim image, GLOW first localizes victim objects according to target labels. And then it generates multiple attack plans, together with their context-consistency scores. Our proposed GLOW, on the one hand, is capable of handling various types of requests, including single or multiple victim objects, with or without specified victim objects. On the other hand, it produces a consistency score for each attack plan, reflecting the overall contextual consistency that both semantic category and global scene layout are considered. In experiment, we design multiple types of attack requests and validate our ideas on MS COCO and Pascal. Extensive experimental results demonstrate that we can achieve about 30$\%$ average relative improvement compared to state-of-the-art methods in conventional single object attack request; Moreover, our method outperforms SOTAs significantly on more generic attack requests by about 20$\%$ in average; Finally, our method produces superior performance under challenging zero-query black-box setting, or 20$\%$ better than SOTAs. Our code, model and attack requests would be made available.
\end{abstract}

\input{tex/intro}
\input{tex/related_works}
\input{tex/method}
\input{tex/experiment}
\input{tex/conclusion}
{\small
\bibliographystyle{ieee_fullname}
\bibliography{egbib}
}

\end{document}

%% file: tex/intro.tex
\section{Introduction}~\label{sec:intro}
Object detection aims to localize and recognise multiple objects in given images with their 2D bounding boxes and corresponding semantic categories~\cite{dalal2005histograms,felzenszwalb2008}. Due to the physical commonsense and viewpoint preferences~\cite{divvala2009empirical}, detected bounding boxes in natural images are not only semantically labeled but also placed relative to each other within a coherent scene geometry, reflecting the 
underlying 3D scene structure. Such bounding box representation allows us to derive a notion of both semantic and geometric constraints. For example, co-occurrence matrix is a commonly exploited semantic constraint where certain object categories are more likely to co-occur, e.g., bed and pillow~\cite{galleguillos2008object}. Geometric constraints, on the other hand, leverage the inductive bias of scene layout~\cite{chen2017spatial}, such as when oc-occurring in a scene, traffic light is more likely to be appeared on the upper region with a smaller bounding box compared to car. 


\begin{figure}[t!]
 \centering
  \includegraphics[width=1.0\linewidth]{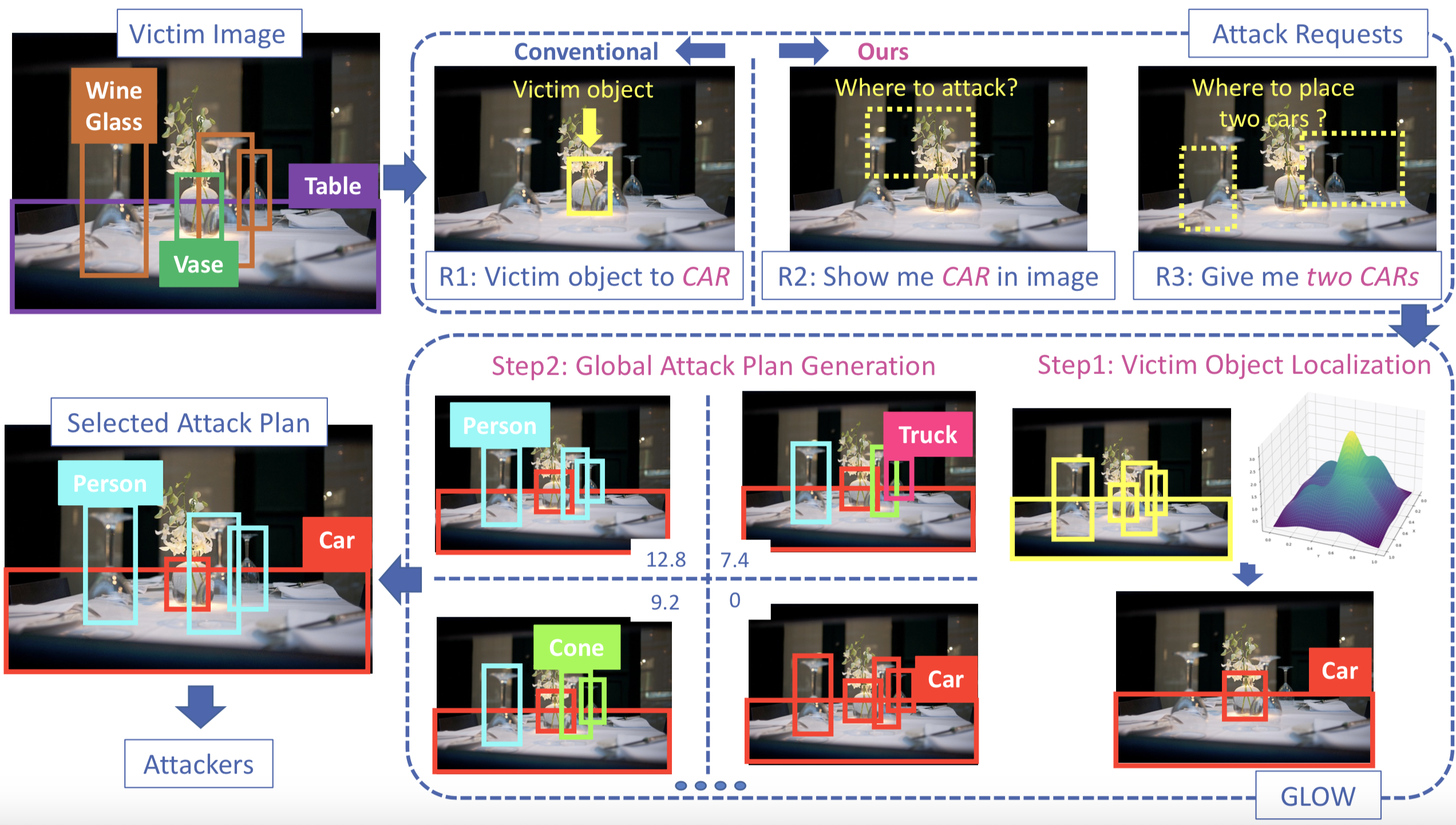}
  \caption{We propose a novel attack generation algorithm GLOW that manages both conventional single targeted object (R1) and our generic attack requests (R2,R3). Specifically, GLOW consists of two steps. The first step localizes victim objects, if not provided. The second step generates various attack plans with their consistency scores. Then the one with the highest score is our final attack plan and parsed to attackers. Best viewed in color.}
  \label{fig:teasor}
\end{figure}

Adversarial attacks on object detectors mainly focus on targeted victim setting~\cite{xie2017adversarial,cai2022context} where the goal is to perturb a specific victim object to target class. In this case, the location, ground truth and target class of the victim object are assumed to be known to attackers. Naturally, contextual cues are leveraged in attack and defense mechanisms~\cite{yin2021exploiting,cai2022context,cai2022zero} on detectors to enhance or detect holistic context (in)consistency~\cite{cai2022zero}.
Though being well-motivated and demonstrating good performances in conventional setting, the state-of-the-art methods~\cite{cai2022context,cai2022zero} suffer the following problems in practice. Firstly, the assumption of known location and ground truth label of victim object might be strong due to annotation cost~\cite{bearman2016s}. Therefore more vague attack requests where victim objects are not specified, e.g. show me an apple and a chair, should be considered in practice, which are beyond the existing methods. Secondly, global geometric layout is commonly neglected as existing methods either model semantic co-occurrence~\cite{cai2022zero} or consider relative sizes and distance w.r.t. given victim object~\cite{cai2022context}.

In this work, we introduce a novel yet generic attack plan generation algorithm GLOW on both conventional and generic attack requests to effectively leverage both categorical and global layout cues, leading to superior white-box attack performance and better transferability in black-box setting. 
As for generic requests, we firstly loose the assumption of known specific victim object by requesting only the existence of certain target label, e.g. show me category X in image. Compared to conventional setting, our request demands the modelling of the locations and sizes of target label X. Our second request further constrains label amount, e.g. give me N objects of category X and M objects of category Y, which necessitates the global layout of victim image.  
To fulfill these requests, we propose a novel attack plan generation method GLOW that accounts for both categorical and geometrical relations. Specifically, GLOW aims to figure out the most context-consistent attack plan for each victim image according to its underlying layout while considering the hard constraints, e.g. existence or amount of some target labels under generic requests or a specific victim object under conventional request. 
The first step in GLOW localizes victim objects with given target label or amount on victim image by modeling the joint distribution of bounding box sizes and centers. And it enables generic attack requests. Given these victim objects, the second step further leverages the layouts of victim images to generate globally context-consistent attack plans with consistency scores. This is achieved by reformulating the generation as a layout similarity measurement problem. And these consistency scores therefore are similarity scores. Finally, the plan with the highest score would be our selected attack plan. We then implement the selected plan with existing attack generation methods, or attackers. Details of our proposed requests and GLOW can be found in Fig.~\ref{fig:teasor}. 


We validate our ideas on coco2017val~\cite{lin2014microsoft} as well as Pascal~\cite{Everingham15} with both white-box and zero-query black-box settings. And we design new evaluation metrics to measure layout consistency thus mimicking consistency defenses. We demonstrate in white-box setting, our proposed method achieves superior performance with both conventional and proposed generic attack setting compared to SOTAs. 
More importantly, GLOW provides significantly better transfer success rates on zero-query black-box setting compared to existing methods.

\noindent Our contributions can be summarized as follows:
\begin{packed_lefty_item}
\item A novel method GLOW that is capable of generating context consistent attack plans while accounts for both categorical and geometric layout coherency.

\item Two generic attack requests on coco2017val and Pascal images and one consistency evaluation metric to mimic realistic attack request and delicate attack defenses.

\item State-of-the-art performances on coco2017val and Pascal images under both white-box and zero-query black-box setting. Code, model and requests will be available.
\end{packed_lefty_item}

%% file: tex/related_works.tex
\section{Related Work}~\label{sec:works}
~\hspace{-7mm}~\noindent{\textbf{Object detection}} The goal of object detection is to predict a set of bounding boxes and category labels for each object of interest. Starting from~\cite{dalal2005histograms,felzenszwalb2008}, object detection explored extensive cues, including semantic~\cite{ladicky2010and}, geometric~\cite{yang2010layered} and other contextual cues~\cite{yao2012describing}, to improve its performance as well as interpretability. Recently, deep neural networks (DNNs)~\cite{krizhevsky2017imagenet} have significantly improved many computer vision~\cite{krizhevsky2017imagenet,girshick2015fast} and natural language processing tasks~\cite{hinton2012deep,devlin2014fast}. Modern detectors follow the neural networks design, such as two-stage models where proposals are firstly generated and then regression and classification are performed~\cite{girshick2015fast,cai2019cascade} and one-stage models~\cite{lin2017focal,zhou2019objects,tian2019fcos} that simultaneously predict multiple bounding boxes and class probabilities with the help of pre-defined anchors or object centers. More recently, transformer-based models~\cite{carion2020end,zhu2020deformable} are proposed to further simplify the detection process by formulating the object detection as a set prediction problem where unique predictions can be achieved by bi-partite matching, rather than non-maximum suppression~\cite{hosang2017learning,bodla2017soft}. Similarly, contextual cues are also explored in modern detectors~\cite{bell2016inside,zhang2017relationship,chen2018context,liu2018structure,barnea2019exploring} with various forms. In this paper, we focus on adversarial attacks on DNNs-based detectors. And our GLOW generates contextually coherent attack plans with various requests, which are also transferable to detectors of different architectures.

\noindent{\textbf{Adversarial attacks and defenses in object detection}} 
Despite impressive performance boosts, DNNs are vulnerable to adversarial attacks, e.g. adding visually imperceptible perturbations to images leads to significant performance drop~\cite{goodfellow2014explaining,szegedy2013intriguing,carlini2017adversarial}. Adversarial attacks can be categorized into white-box~\cite{goodfellow2014explaining,madry2017towards} and black-box~\cite{dong2018boosting,lin2019nesterov}, depending on whether parameters of victim models are accessible or not.
Attacks such as DAG~\cite{xie2017adversarial}, RAP~\cite{li2018robust} and CAP~\cite{zhang2020contextual} are architecture-specific white-box attacks on detectors where two-stage architecture is required since they work on proposals generated by the first stage. More generic attacks, such as UAE~\cite{wei2018transferable} and TOG~\cite{chow2020adversarial}, are capable of attacking all different kinds of models regardless of their architectures. 
Compared to the aforementioned methods that perturb the image globally~\cite{xie2017adversarial}, patch-based attacks~\cite{liu2018dpatch} also showcase their ability in terms of fooling the detectors without touching the victim objects~\cite{hu2021cca}. In contrast, black-box attacks~\cite{li2020practical,chakraborty2018adversarial,papernot2017practical,liu2016delving,li2020towards} are more practical yet challenging where either a few queries or known surrogate models are exploited to fool an unknown victim model.
Observing the impacts of adversarial attacks on detectors, various defense methods are proposed to detect such attacks, wherein contextual cues are explored~\cite{yin2021exploiting}. However, contextual cues are almost always represented in the form of semantic co-occurrence matrix where global layouts are largely neglected~\cite{cai2022context,cai2022zero}.
In contrast, we propose a generic attack plan generation algorithm that leverages both semantic and geometric coherency, e.g. scene layout. Consequently, it manages both conventional single targeted victim setting and generic attack requests where locations are unknown or object amount is further restricted, translating to SOTA performance under white-box and black-box settings.


%% file: tex/method.tex
\section{Method}~\label{sec:method}
We introduce the attack requests, proposed GLOW and attacker in Sec.~\ref{sec:attk_req}, Sec.~\ref{sec:generation} and Sec.~\ref{sec:imp_ap} respectively. 

\subsection{Attack requests}~\label{sec:attk_req}
To attack a victim image, user may or may not specify victim objects, e.g. providing their locations or labels. Therefore, besides considering conventional attack request where a specific object and its targeted label are given, more generic requests should also be addressed, 
such as give me 2 cats or mis-classify the rightmost boat to car. 
Let's denote $\mathcal{D}$ as the set of multiple victim image $I$.
$\mathcal{C}=\{c_p\}_{p=1}^C$ is the label space with $C$ semantic categories. Given a known object detector $f$, which can be the victim model in white-box attack or the surrogate model under black-box setting, we can obtain a set of predicted objects $\mathcal{O}$ on victim image $I$, or $\mathcal{O}=f(I)=\{l_n, s_n\}_{n=1}^N$, consisting the location and semantic category of $N$ objects. 
$l_n$ defines the location of the $n$-th object, including its bounding box center coordinates, height and width.
And $s_n\in\mathcal{C}$ is its semantic label.

\noindent{\textbf{R1: mis-classify the object $s_n$ to $c_p$.}} This is the conventional attack request where the $n$-th object is our specific victim object and $c_p$ is the targeted label.

Though one can always choose random object as victim and random category as $c_p$, we observe that the choice of victim object and target label plays an important role in attack performances (See Sec.~\ref{sec:exps}). 
To this end, we set different selection criteria for victim object and targeted label to evaluate attack methods in various aspects. As for victim object, it is unpractical to assume that ground-truth locations can be provided by the users, e.g. bounding box annotations can be time-consuming~\cite{bearman2016s}. Therefore, we turn to the predictions as reliable sources to help us to determine where the attack should take place. More specifically, the one that has the largest bounding box among all predictions with confidence score above 0.85 will be selected, which gives us good estimation in practice. 

As for target label $c_p$, we mainly follow~\cite{cai2022context,cai2022zero} where the out-of-context attack is considered. Specifically, to eliminate the chance of miscounting the existing objects as success, $c_p$ is selected if and only if $c_p$ is not present in the $I$. Rather than randomly selecting $c_p$ among all unpresented categories~\cite{cai2022context,cai2022zero}, our decision is made according to distance in word vector space~\cite{yu2019mcan} as it captures the semantic and syntactic qualities of words. Mathematically, for each unpresented $c_p$, we have its averaged distance as:
\begin{equation}
    v_d(c_p)=\frac{1}{N}\sum_{n}v(c_p,s_n); \forall c_p\notin\mathcal{S}_n
\end{equation}~\label{eq:label_dis}
where $\mathcal{S}_n=\{s_n\}_{n=1}^N$ and $v(c_p,s_n)$ denotes the cosine distance between category $c_p$ and $s_n$ in word vector space. 

\begin{figure}[t!]
 \centering
  \includegraphics[width=1.0\linewidth]{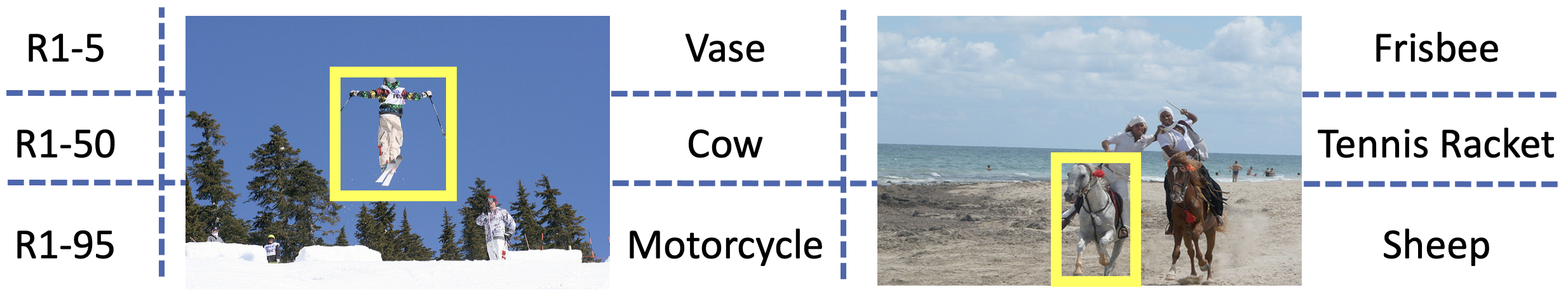}
  \caption{ Examples of R1. Victim objects $s_n$ are highlighted with yellow bounding boxes and target labels $c_p$ generated with R1-5, R1-50 and R1-95 are on the right side of each victim image.
  }
  \label{fig:r1_example}
\end{figure}

To evaluate the impact of target label $c_p$, we collect three $c_p$s according to $v_d(c_p)$ and visualize them in Fig.~\ref{fig:r1_example}. Specifically, we firstly rank all $c_p\notin\mathcal{S}_n$ based on $v_d(c_p)$. Then we choose the top $5\%$, $50\%$ and $95\%$ ones as our target class $c_p$s, referring as R1-95, R1-50 and R1-5, respectively.

Our ultimate goal of R1 is not only to mis-classify the victim object, but also failing the potential defense of consistency check. Therefore, the challenge of R1 mainly lies in figuring out the attack plan that is contextually consistent and beneficial for the mis-classification in practice. 

\noindent{\textbf{R2: show me the category $c_p$.}} Rather than assuming that a specific victim object is known to attackers as in R1, R2 takes one step further in terms of relaxing the attack request. Specifically, R2 comes in a much vague manner where user only specifies the target label $c_p$. 

Though it seems that asking for the existence of $c_p$ is an easier task compared to R1 as one can always flip a random object to $c_p$, we argue that this conclusion is valid if only coarse semantic consistency check/defense, e.g. co-occurrence matrix~\cite{cai2022zero}, is available,  which unfortunately neglects geometric context.
A more desired consistency check should be capable of capturing both geometric and semantic context, 
for example, traffic light is less likely to appear on the image bottom while poles usually has slim bounding boxes. And our goal is to fool the victim model and such delicate defense simultaneously. 

\begin{figure}[t!]
 \centering
  \includegraphics[width=1.0\linewidth]{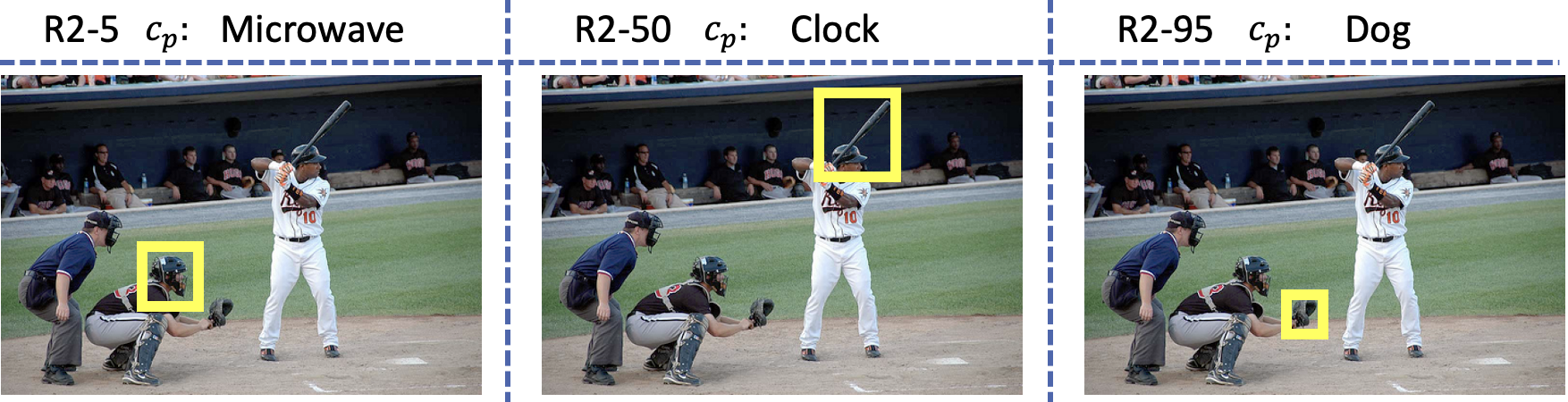}
  \caption{We visualize one victim image with their R2-5, R2-50 and R2-90 labels. And we highlight the victim object $s_n$, which is localized by GLOW, with yellow bounding box. 
  }
  \label{fig:r2_example}
\end{figure}

Therefore, we claim that R2 is more challenging than R1 as it requests additional understanding of the location-wise distribution of target label $c_p$.
We kindly note our readers that such challenge is beyond~\cite{cai2022zero,cai2022context} (see Fig.~\ref{fig:r2_example}). We omit the details of $c_p$ in R2-5, R2-50 and R2-95 as they are selected based on the same criteria as that of R1 (See supplementary).

\begin{figure}[t!]
 \centering
  \includegraphics[width=1.0\linewidth]{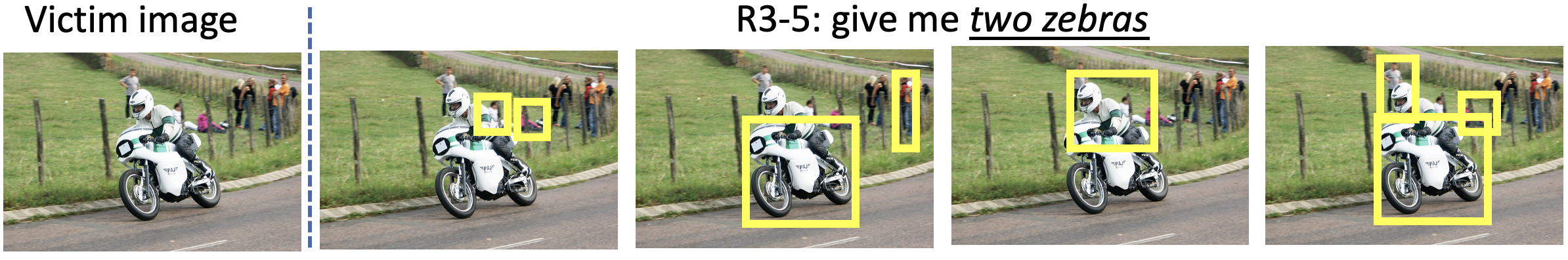}
  \caption{Challenge of R3. We have victim image on the left and four example proposals based on request R3-5 on the right. Among these four proposals, the left most one is more plausible than one in middle considering the layout relations. And the right two are totally wrong as they violate the amount restriction.}
  \label{fig:r3_example}
\end{figure}

\noindent{\textbf{R3: give me multiple $c_p$s.}} R3 reflects another realistic attack scenario, e.g. have a monitor and a mouse in victim image $I$. Besides not specifying the victim object by providing only target label information, 
R3 enforces additional constraint on object amount, making it more challenging. Specifically, multi-object relationship should be considered together with hard restrictions on the amount of objects (See Fig.~\ref{fig:r3_example}). For example, besides modelling locations of mouse and monitor individually, estimating their layout, e.g. monitor is more likely to be above the mouse, is also essential to achieve context consistent yet fooled predictions. 

Theoretically, R3 can be multiple victim objects of the same or different categories, which does not affect our following GLOW method. In practice, when it comes to objects with various categories, additional heuristics are needed to avoid semantic inconsistency as $v_d$ does not guarantee contextual consistent combinations. Moreover, such problem becomes more severe with increasing number of objects, together with the emerge of new challenge of underlying constraints on object amount in natural image. For instance, ten~\textit{apples} in $I$ can be natural but not for ten~\textit{stop signs}. 
Therefore, we leave objects of different categories as our future work and focus on two objects of the same $c_p$. Details of R3-5, R3-50 and R3-95 can be found in supplementary.


\begin{figure*}[t!]
 \centering
  \includegraphics[width=1.0\linewidth]{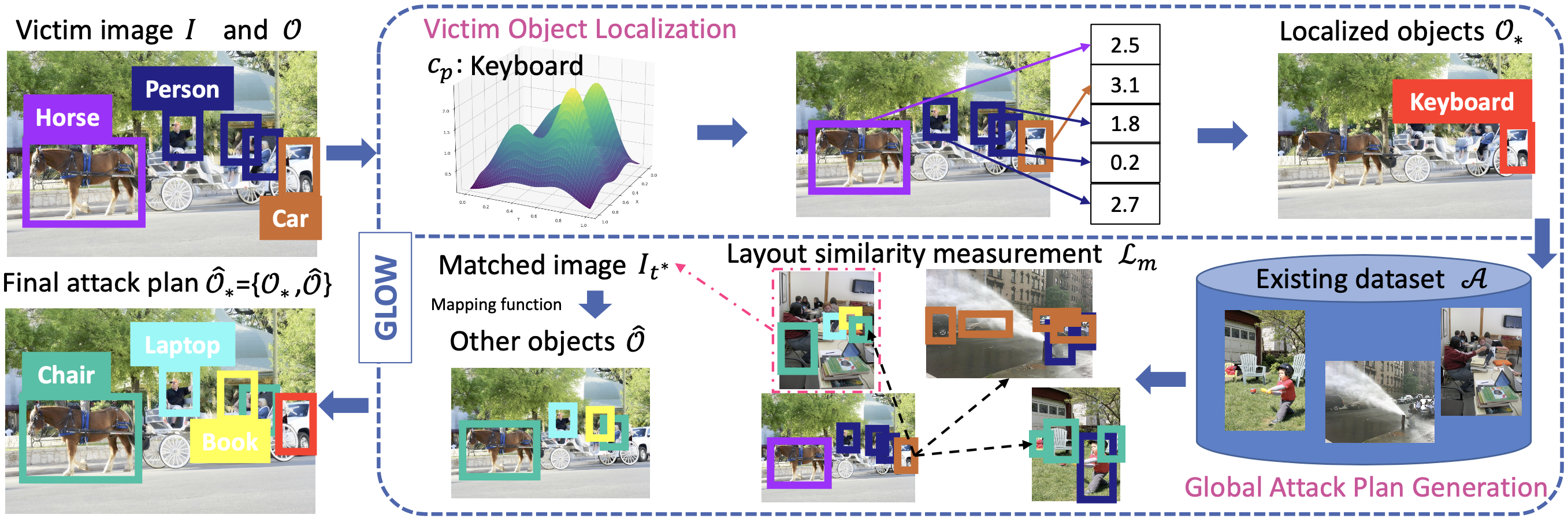}
  \caption{Overview of GLOW. The first step of GLOW aims to locate the victim object $\mathcal{O}_*$ under generic attack requests according to dataset distribution. 
  Afterwards, the GLOW produces various context-consistent attack plans, together with their consistency scores. 
  The plan with highest score is selected as our final attack plan $\hat{\mathcal{O}}_*$.}
  \label{fig:main}
\end{figure*}

\subsection{GLOW: Global LayOut aWare attacks}~\label{sec:generation}
Contextually consistent attack has been discussed in many previous work~\cite{cai2022context,cai2022zero}. The main motivation is that perturbing only the victim object may lead to inconsistency in context thus global attack plan should be considered. Specifically, an attack plan assigns target labels to all objects in victim image, including ones that are not victims originally, to both avoid inconsistency and benefit the attack request. Though well-motivated, existing methods largely rely on semantic context~\cite{cai2022zero},
neglecting geometric context such as scene layout. 
In addition, the ability of modelling prior knowledge, such as having more than ten beds in an image is unlikely to happen while ten books are more plausible, is lacking in literature.

To this end, we propose a novel attack plan generation method GLOW that accounts for both semantic and geometric context, such as object locations and overall scene layout. GLOW consists of two steps. The first localization step aims to locate the victim object based on target labels and their amounts under generic attack requests R2 and R3. Then the second generation step further produces multiple context-consistent attack plans as well as their scores with given victim objects. 
Afterwards, the plan with the highest score is selected as our final attack plan and then parsed to existing attackers. See Fig.~\ref{fig:main} for more details.

\noindent{\textbf{Victim object localization}} We aim to localize victim objects under R2 and R3, where constraints on target labels and/or their amount are available. 

Let's first assume there exist some images from annotated detection dataset, which, in the simplest case, can be the training set that our victim/surrogate model is trained on. 
We denote this dataset as $\mathcal{T}$, including $T$ images and their bounding box annotations $\mathcal{A}=\{\mathcal{A}_t\}_{t=1}^T$, where $\mathcal{A}_t=\{l^t_m,s^t_m\}_{m=1}^M$ is the set of bounding box annotations on the $t$-th image $I_t$. Similarly, we assume there exist $M$ objects and the $m$-th object instance has location $l^t_m$ and semantic category $s^t_m\in\mathcal{C}$.


Determining the location of victim object under R2 and R3 is equivalent to estimating the center, height and width of bounding boxes of target label $c_p$. And we formulate the localization as a probability maximization problem. This is achieved by modelling the joint probability of bounding box center, height and width per category.
Specifically, for each $c_p\in\mathcal{C}$, we have $\mathcal{L}_{c_p}=\{l^t_m|s^t_m=c_p\}_{t,m}$, where $l^t_m$ is normalized by image height and width. Then we apply GMM~\cite{rasmussen1999infinite} to fit $q=\{1,\dots,Q\}$ Gaussians $\mathcal{N}^p_q(\mu^p_q,\delta^p_q)$ on $\mathcal{L}_{c_p}$, where $\mu^p_q$ and $\delta^p_q$ are mean and co-variance of $q$-th Gaussian at class $c_p$. $pdf_q^p$ and $\pi^p_q$ are the probability density function and the weight of $\mathcal{N}^p_q$ respectively. Q is set to 5 based on experiment on$~\mathcal{T}$. 
Given any $x\in\mathbb{R}^4$,
our GMM is able to provide a weighted probability density $w(x)$ by:
\begin{equation}
    w_p(x)=\frac{1}{Q}\sum_{q}\pi^p_q\times pdf_q^p(x)
\end{equation}

\begin{figure*}[t!]
 \centering
  \includegraphics[width=1.0\linewidth]{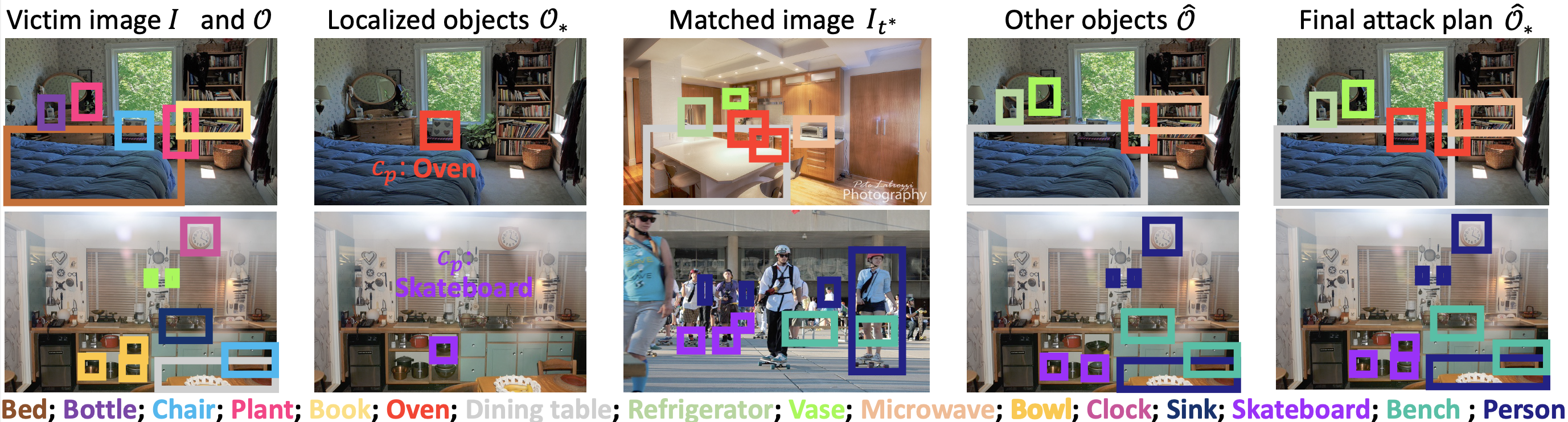}
  \caption{Step-wise illustration of GLOW. From left to right, we have victim image $I$ with their initial prediction results $\mathcal{O}$, target label $c_p$ and the localized victim object $\mathcal{O}_*$, best matching image $I_{t^*}$, the plan for other objects $\hat{\mathcal{O}}$ and our final attack plan $\hat{\mathcal{O}}_*$.}
  \label{fig:process}
\end{figure*}

Simply going through all $x$ and choosing ones with highest $w_p(x)$ ignore overall scene layout, which might result in significant layout changes, e.g. large bounding box on objectless area or heavy occlusions, leading to less plausible overall layouts.
Alternatively, we narrow down our search space to existing bounding boxes and find the optimal location among all $l_n$. As for R2, the victim object can be found by $n^*=\arg\max_n w_p(l_n)$. As for R3, we rank and select top ones depending on detailed request, e.g. choose the top 2 if R3 is to have two objects of same target label $c_p$. We then denote the victim objects in $I$ as $\mathcal{O}_{*}=\{l_p^*,c_p^*\}_*$, where $c_p^*$ equals to $c_p$ in R1 and R2. And $\{c_p^*\}_*$ is the set of requested target labels in R3. Similarly, $l_p^*$ is $l_n$ in R1 and are from estimation in R2 ($l_{n^*}$) and R3. We further denote the number of target objects as $X$ and $X=1$ under R1 and R2. 
Example victim objects $\mathcal{O}_*$ can be found in Fig.~\ref{fig:process}. 


\noindent{\textbf{Global attack plan generation}} Given victim object $\mathcal{O}_{*}$, our next step is to generate target labels on objects that are not victim. 
Specifically, it aims to find an mapping function $g(s_n)\in\mathcal{C}$ that perturbs the label of these objects, resulting in $\hat{\mathcal{O}}=\{l_n,g(s_n)\}_n$. The overall generated attack plan on $I$ would be $\hat{\mathcal{O}}_*=\{\mathcal{O}_*,\hat{\mathcal{O}}\}$.

Theoretically, there exist $(N-X)^C$ possible configurations in $\hat{\mathcal{O}}$. Instead of permuting all possible solutions, we restrict ourselves with only feasible ones that occur in existing dataset $\mathcal{T}$ as scene layouts are naturally context-consistent therein. To this end, we formulate our global attack plan generation as a layout similarity measurement problem, with hard constraint on victim objects $\mathcal{O}_*$. Our goal is therefore to map the bounding box labels according to the best match based on layout similarity in $\mathcal{T}$. Intuitively, the more similar these layouts are, the more confident we are in terms of performing mapping. Therefore the layout similarity score reflects context consistency to some extent. Our insights lie in the following design choice of obtaining mapping function $g$ and score $s$:
\begin{packed_lefty_item}
    \item Generate $\mathcal{T}^*=\{\mathcal{T}_{c_p^*}\}_{c_p^*}$ where $\mathcal{T}_{c_p^*}$ consists of images that have target label $c_p^*$ presented. 
    \item Compute the Intersection over Union (IoU) score between victim objects $\mathcal{O}_*$ and objects that share the same target labels in $I_t\in\mathcal{T}_{c_p^*}$, 
    Mathematically,:
    \begin{equation}
        s_1(I_t)=\frac{1}{X}\sum_{l_p^*\in\mathcal{O}_*}s(l_p^*); \forall I_t\in\mathcal{T}^*
    \end{equation}
    where $s(l_p^*)=\max_m \mathbbm{1}_{\{c_p^*=s^t_m\}}IoU(l_p^*,l^t_m)$. The IoU score between victim object location $l_p^*$ and $m$-th bounding box in $I_t$ is obtained by $IoU(l_p^*,l^t_m)$. 
    \item Perform Hungarian matching~\cite{stewart2016end,carion2020end} between objects in $I_t$ and those in victim image $I$. 
    Specifically, we find a bipartite matching between these two sets by searching for a permutation of $M$ elements $\mathbb{S}_M$ with the lowest cost:
    \begin{equation}
    \begin{split}
        \delta_t^* & = \arg\max_{\delta\in\mathbb{S}_M}\sum_{l_n\notin\mathcal{O}_*} \mathcal{L}_m(l_n,l^t_{\delta(n)}) \\
        & = \arg\max_{\delta\in\mathbb{S}_M}\sum_{l_n\notin\mathcal{O}_*} L1(l_n,l^t_{\delta(n)})+GIoU(l_n,l^t_{\delta(n)})
    \end{split}
    \end{equation}
    where $L1()$ and $GIoU()$ define the L1 and GIoU~\cite{rezatofighi2019generalized} between bounding boxes. $\delta_t^*(n)$ is the index of the best match of $n$-th object which is not victim originally in victim image $I$. And the match loss of $\delta_t^*(n)$ can be obtained with $s2(I_t)=\frac{1}{N-X}\sum_{l_n\notin\mathcal{O}_*}\mathcal{L}_m(l_n,l^t_{\delta_t^*(n)})$. The temporary mapping function based on the $t$-th image $I_t$ is then defined as $g_t(s_n)=s^t_{\delta_t^*(n)}$.
\end{packed_lefty_item}
The overall similarity score between $I_t\in\mathcal{T}^*$ and $I$ is obtained by $s(I_t)=s_1(I_t)-\lambda s_2(I_t)$, 
where $\lambda$ is a hyper-parameter chosen by experiment. We would like to note that score $s$ accounts for not only the victim objects reflecting by $s_1$, but also the overall layout similarity incorporated in $s_2$. 

Afterwards, we find the $I_{t^*}$ as long as it 1) gives the highest similarity score and 2) matches more than 95$\%$ of objects in $I$. 
Consequently, the mapping function $g(s_n)$ then equals to the temporary mapping function of the $t^*$-th image, or $g_{t^*}(s_n)$. 
We refer the readers to Fig.~\ref{fig:process} for more details.



\subsection{Implementation of attack plan}~\label{sec:imp_ap}
To generate $\hat{\mathcal{O}}_*$, evasion attacks can be implemented using our victim model itself under white-box setting or a single or multiple surrogate model(s) under zero-query black-box setting. In white-box scenario, our implementation of attack plan is based on TOG~\cite{chow2020adversarial} for fair comparisons with existing methods~\cite{cai2022context}(see Sec.~\ref{sec:exps}). Specifically, we fix the weight of victim model $f$ and learns a perturbation image $\delta$ for $I$ by minimizing $\mathbf{L}(clip(I+\delta);\hat{\mathcal{O}}_*)$ at every iteration~\cite{goodfellow2014explaining}. $clip()$ is enforced to ensure bounded perturbation.
Afterwards, the perturbed image $clip(I+\delta)$ is parsed to another unknown victim model, mimicking the zero-query black-box setting. 

%% file: tex/experiment.tex
\section{Experiment}~\label{sec:exps}
To evaluate GLOW under various requests, we perform extensive experiments on coco2017val~\cite{lin2014microsoft} and Pascal~\cite{Everingham15}, with both white-box and black-box settings. As can be found in Tab.~\ref{tbl:exp_setup}, our victim model $f$ can be Faster-RCNN-R50-FPN-1X-COCO($\mathbb{F}$)~\cite{ren2015faster} and F-RCNN+YOLO($\mathbb{F}$+$\mathbb{Y}$)~\cite{redmon2016you} under white-box setting. These aforementioned victim models are later utilized as the surrogate model in our black-box attacks where 
DETR($\mathbb{D}$)~\cite{carion2020end} and RetinaNet($\mathbb{T}$)~\cite{lin2017focal} are our victim models $f$. Our black-box attack is zero-query based, meaning no feedback from victim model is available. Our GLOW is generally applicable to different victim detectors and we choose the aforementioned models mainly for efficiency and re-productivity purpose~\cite{cai2022context}. We report our performance under both perturbation budget 10 and 30. Due to the space limitation, we refer the readers to supplementary materials for results with the former and discussions on limitations. And our claims are valid with different perturbation budgets. 


\begin{table}[t!]
    \centering
    \resizebox{\columnwidth}{!}{
    \input{tables/exp_setup}}
    \caption{Experimental setup. Our victim model is trained on coco17train only and the victim images are from coco17val and Pascal. The former consists of 3792 images with 80 categories while the latter has 500 images of 20 categories. 
    }
    \label{tbl:exp_setup}
\end{table}



\begin{table*}[t!]\small
    \centering
    \resizebox{\textwidth}{!}{
    \input{tables/res_r1_iccv_v1}}
    \caption{Overall performance of R1. As described in Sec.~\ref{sec:attk_req}, we have three different target labels, R1-5, R1-50 and R1-95, for each victim object and we report the results on all of them. We highlight the best and second best with bold and underline respectively. 
    }
    \label{tbl:r1_res}
\end{table*}


\noindent{\textbf{Baselines}} We compare GLOW with four baselines. To perform fair comparison, attack plan implementations are all obtained with TOG~\cite{chow2020adversarial} thus we describe only the attack plan generation process in the following:
\begin{packed_lefty_item}
    \item TOG~\cite{chow2020adversarial} The attack plan generated by the TOG is context-agnostic, or $g(s_n)=s_n$. 
    Victim object is given in R1 and will be randomly selected under R2 and R3.
    \item TOG+RAND. TOG+RAND. focuses on both victim objects and other objects. Victim object is provided in R1 and randomly selected under R2 and R3. 
    Mapping function $g(s_n)$ is a random permutation function $r$. 
    \item TOG+SAME. Attack plan generated by TOG+SAME. includes all objects. 
    And we enforce $g(s_n)=c_p$, meaning all objects share the same target label $c_p$. 
    \item Cai~\cite{cai2022context} Cai~\cite{cai2022context} can be directly apply to R1. As for R2 and R3, Cai~\cite{cai2022context} firstly selects random objects as victims and then generates the attack plan.
\end{packed_lefty_item}

\begin{figure}[t!]
 \centering
  \includegraphics[width=1.0\linewidth]{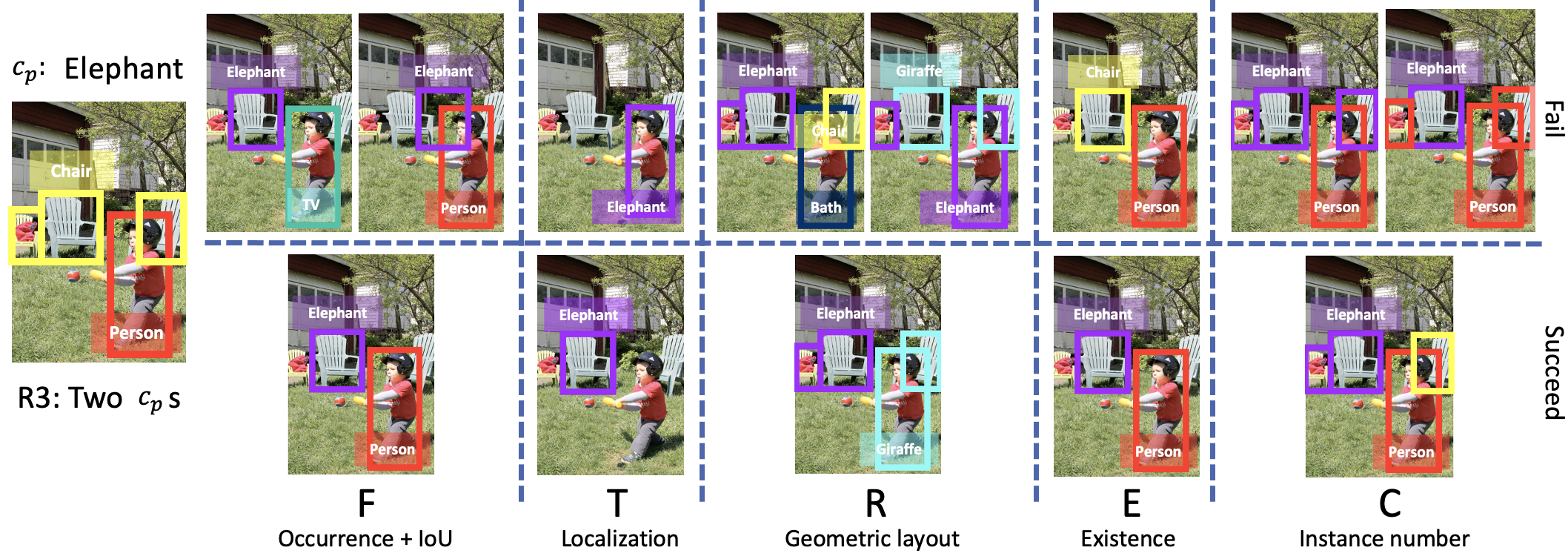}
  \caption{We visualize failure and successful cases of different evaluation metrics and their key factors under R3, where our goal is to have two elephants in given victim image.}
  \label{fig:eval_exp}
\end{figure}

\noindent{\textbf{Evaluation Metrics}} We follow the basic metric from~\cite{cai2022zero} and also introduce others for generic attack requests. Fooling rate ($\textbf{F}$)~\cite{cai2022zero} is used to evaluate the attack performance on victim objects. Specifically, one attack succeeds if (1) victim object is perturbed as target label while IOU is score greater than 0.3 compared to GT and (2) it pass the co-occurrence check. And we define the fooling rate as the percentage of the number of test cases for which the above two conditions are satisfied. Besides, we further introduce $\textbf{T}$ to measure the consistency on victim objects. $\textbf{T}$ itself reveals the averaged $w_p(l_p^*)$. When combined with other metrics, $\textbf{T}$ is satisfied as long as the averaged $w_p(l_p^*)$ is above 0.02 (see Sec.~\ref{sec:generation}).
To measure the overall layout consistency, we introduce $\textbf{R}$ that reflects the percentage of images whose maximum recall rate compared to $\mathcal{A}$ is above 0.5.
We further design two metrics, $\textbf{E}$ and $\textbf{C}$, on R2 and R3 to report successful rate. $\textbf{E}$ checks whether target label $c_p$ exists in predictions. While $\textbf{C}$ further verify the amount of $c_p$. One attack is successful if both target labels and their amount satisfy the request in R3. We refer the readers to supplementary for more details of all metrics and give some visual examples in Fig.~\ref{fig:eval_exp}.

\subsection{Main results}~\label{sec:white_main}
~\hspace{-7mm}\noindent{\textbf{Attack performance on R1}} We report our main results on conventional attack request R1 in Tab.~\ref{tbl:r1_res} where perturbation budgets is set to 30. In general, we observe that under white-box setting, our $\textbf{F}$ is comparable to existing methods on coco, which is reasonable as this metric considers only oc-occurrence matrix and both TOG+SAME and Cai~\cite{cai2022context} considers this semantic consistency. 
When considering global layout $\textbf{R}$, we observe clear performance improvement, or 30$\%$, over existing methods under all scenarios (R1-5, R1-50, R1-95), meaning that GLOW is able to not only fool the victim object, but also give more contextually consistent layout. 
Noticeably, our observation of 30$\%$ averaged improvement is also valid under challenging zero-query black-box setting, which further demonstrates the transferbility of our proposed attack plan generation. Please note that results on Pascal are obtained with victim models that trained on coco17train, which showcases the generality and superiority of GLOW. 

There are also other interesting observations in Tab.~\ref{tbl:r1_res}. Firstly, there exists a trend of performance improvement over all methods when compared R1-5, R1-50 and R1-95, indicating the selection of target label plays an important role in terms of performance. This trend validates our hypothesis that far-away labels, e.g. R1-5, are harder to attack compared to close-by ones, which in return proves the necessity of systematic design on target label rather than random generation. Secondly,  
though TOG+SAME simply assigns all labels of existing objects to be target label $c_p$, it gives good performance under $\textbf{F}$. This observation further supports our design of more delicate consistency check metrics, e.g. $\textbf{R}$, as co-occurrence matrix is vulnerable to such simple hacks.

\begin{table*}[t!]\small
    \centering
    \input{tables/res_r2_iccv_v1}
    \caption{Overall performance of R2. Similar to R1, we have three different target labels for victim image. Since the victim object location is not provided in R2, $\textbf{T}$, $\textbf{F+T}$ and $\textbf{E+R}$ reflects different aspects of layout consistency.}
    \label{tbl:r2_res}
\end{table*}

\begin{table*}[t!]\small
    \centering
    \input{tables/res_r3_iccv_v1}
    \caption{Overall performance of R3. Compared to R2, our $\textbf{C+R}$ accounts for both layout consistency and amount restriction.}
    \label{tbl:r3_res}
\end{table*}

\begin{figure}[t!]
 \centering
  \includegraphics[width=1.0\linewidth]{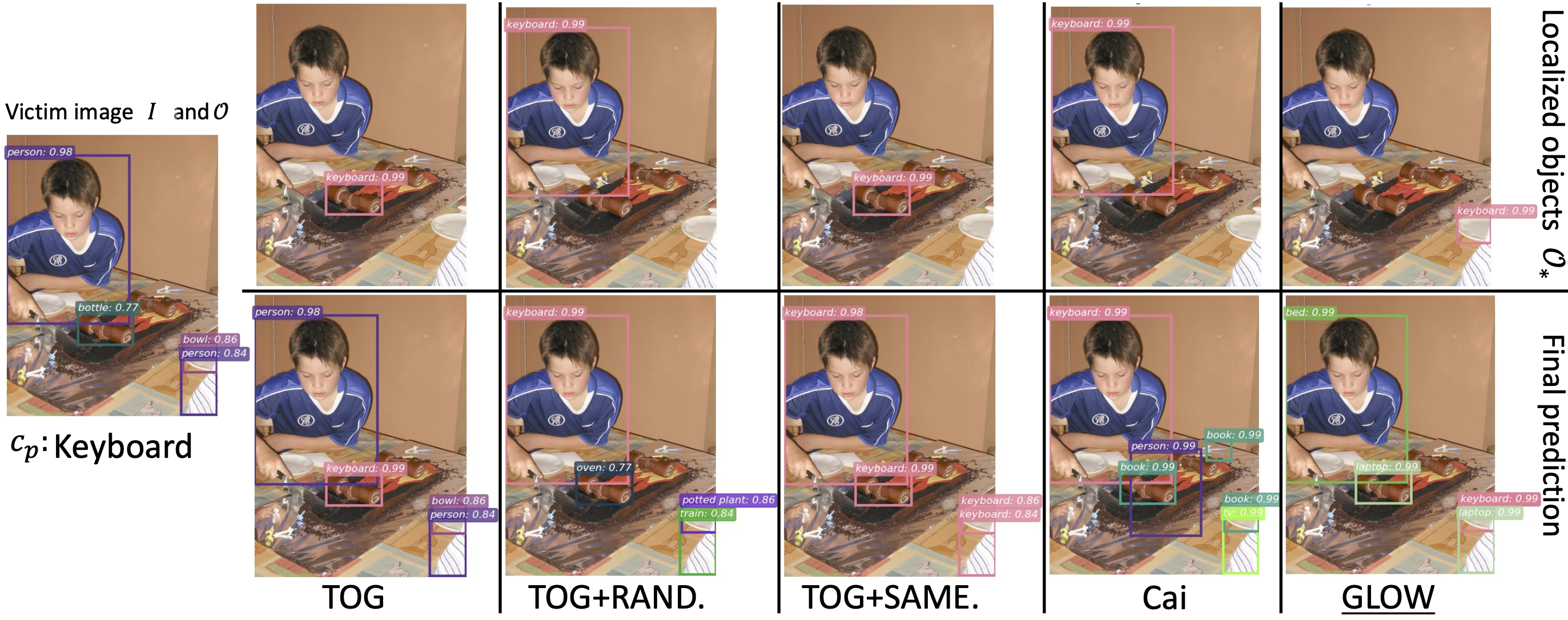}
  \caption{We visualize the results of R2-5 where the $c_p$ is keyboard. GLOW makes reasonable target object localization and context consistent attacks compared to all other baselines.}
  \label{fig:baseline_exp}
\end{figure}

\noindent{\textbf{Attack performance on R2}} 
The advantages of GLOW are more noticeable in R2 where victim object is requested to be localized by algorithm itself rather than being provided, as can be observed in Tab.~\ref{tbl:r2_res}. There are two main observations. Firstly, GLOW almost always beats the SOTAs in terms of all evaluation metrics under R2-5, R2-50 and R2-95 in white-box setting, e.g. about 35$\%$ relative improvement compared to the second best in terms of $\textbf{F+T}$ and $\textbf{E+R}$ under R2-5. This observation is also valid when victim models are trained on coco17train and tested on Pascal. Interestingly, unlike R1 where victim object is fixed among all methods, results of $\textbf{T}$ in R2 showcase that the victim object selection matters under generic request. 
Though neither TOG+SAME nor Cai~\cite{cai2022context} considers the overall layout consistency, the former gives better score compared to the later as it naively enforces all objects share the same target label and \textbf{E} in \textbf{E+R} measures only the existence of target label. Please note that \textbf{E} and \textbf{F} are different.  
For instance, assuming layout consistency is already satisfied, if one attack on the victim object fails but turns another object into target label, it will be regarded as success in \textbf{E} but failure in \textbf{F}. GLOW, again, produces superior results by leveraging layout explicitly. Our second observation from Tab.~\ref{tbl:r2_res} is that GLOW has better transfer rates, such as 24$\%$ improvement compared the context-aware baseline~\cite{cai2022context} and various types of random assignment under black-box setting, which further showcases the benefits of utilizing global layout in attack plan generation and the potential limitation of exploiting only semantic context. We observe the same trend that the overall performance improves when the target label is closer to presented labels in word space, supporting our design of various target label. We visualize some examples and results in Fig.~\ref{fig:baseline_exp}. We observe that GLOW provides most reasonable semantic configurations given current scene layouts compared to all baselines.



\noindent{\textbf{Attack performance on R3}} 
Results of most challenging request R3 are provided in Tab.~\ref{tbl:r3_res}.
We kindly remind our readers that \textbf{C+R} and \textbf{F+T+C} reflect different aspects of an algorithm as the former does not care about specific objects but checks both target labels and their the amount. Assuming our R3 is to have two apples in victim image and our attacks are contextually consistent, \textbf{F+T+C} will be successful if two victim objects are perturbed to apple. 
In contrast, \textbf{C+R} reflects the amount of apples in perturbed images and mismatch in numbers would lead to failure. Again, our GLOW is a much safer choice in terms of R3 as it can almost always, or at least 12 out of all 18 entries, give the best performance with both white-box and black-box setting.






%% file: tables/exp_setup.tex
\begin{tabular}{c|c|c|c|c|c}
    \toprule
    \multirow{2}{*}{Victim $f$} 
    & Whi. & \multicolumn{2}{c|}{$\mathbb{F}$} &\multicolumn{2}{c}{$\mathbb{F}$+$\mathbb{Y}$} \\ \cline{2-6}
    & Blk. & \multicolumn{2}{c|}{$\mathbb{F}\rightarrow\mathbb{D}$} & \multicolumn{2}{c}{$\mathbb{F}$+$\mathbb{Y}\rightarrow \mathbb{T}$}\\ \hline \hline
    \multicolumn{2}{c|}{Victim set$~\mathcal{D}$} 
    & \multicolumn{2}{c|}{coco17val (3792/80)} 
    & \multicolumn{2}{c}{Pascal (500/20)} 
    \\ \hline \hline
    \multicolumn{4}{c|}{Victim model $f$ is trained on and $~\mathcal{T}$} & \multicolumn{2}{c}{coco17train} \\
    \bottomrule
\end{tabular}

%% file: tables/res_r1_iccv_v1.tex
\begin{tabular}{c|cc|cc|cc|cc|cc|cc}
    \toprule
    \multirow{3}{*}{Methods} 
    &\multicolumn{6}{c|}{White-box (coco17val/Pascal)}
    &\multicolumn{6}{c}{Zero query black-box (coco17val/Pascal)} \\ \cline{2-13}
    &\multicolumn{2}{c|}{R1-5}
    &\multicolumn{2}{c|}{R1-50}
    &\multicolumn{2}{c|}{R1-95}
    &\multicolumn{2}{c|}{R1-5}
    &\multicolumn{2}{c|}{R1-50}
    &\multicolumn{2}{c}{R1-95}\\ \cline{2-13}
    &F & F+R &F  & F+R &F & F+R &
    F & F+R &F & F+R &F & F+R \\
    \midrule
    TOG~\cite{chow2020adversarial} & .64/\textbf{.67} & \underline{.11}/\underline{.16} & .75/\textbf{.77} & .15/\underline{.22} & .87/\textbf{.82} & \underline{.20}/\underline{.27} & .08/\textbf{.13} & .01/\underline{.04} & .16/.19 & .02/.08 & .23/.27 & .03/\underline{.13}     \\ 
    TOG+RAND & .45/.48 & .06/.08 & .54/.61 & .08/.14 & .58/.66 & .07/.14 & .12/.10 & .01/.02 & .21/.17 & \textbf{.03}/.04 & .27/.26 & \textbf{.04}/.06   \\ 
    TOG+SAME &  \textbf{.89}/.52 & .18/.13 & \textbf{.90}/.68 & \underline{.18}/.22 & \textbf{.91}/.75 & .18/.22 & \textbf{.21}/.10 & .01/\underline{.04} & \textbf{.34}/\textbf{.22} & \textbf{.03}/\underline{.11} & \textbf{.38}/\underline{.31} & .03/.11   \\ 
    Cai~\cite{cai2022context} & \underline{.86}/.46 & .09/.08 & \underline{.87}/.63 & .09/.09 & \underline{.90}/.74 & .07/.11 & .18/.08 & .01/.03 & .29/.19 & .02/.08 & .34/.30 & .02/.11  \\ 
    \midrule
    GLOW & .85/\underline{.61} & \textbf{.20}/\textbf{.20} & \underline{.87}/\underline{.76} & \textbf{.22}/\textbf{.28} & .89/\underline{.79} & \textbf{.21}/\textbf{.29} & \textbf{.21}/\underline{.11} & \textbf{.02}/\textbf{.05} & \underline{.30}/\textbf{.22} & \textbf{.03}/\textbf{.12} & \underline{.35}/\textbf{.33} & \textbf{.04}/\textbf{.18}  \\
    \bottomrule
\end{tabular}

%% file: tables/res_r2_iccv_v1.tex
\begin{tabular}{c|ccc|ccc|ccc}
    \toprule
    \multirow{3}{*}{Methods} 
    &\multicolumn{9}{c}{White-box (coco17val / Pascal)} \\ \cline{2-10}
    &\multicolumn{3}{c|}{R2-5}
    &\multicolumn{3}{c|}{R2-50}
    &\multicolumn{3}{c}{R2-95} \\ \cline{2-10}
    & T & F+T & E+R & T & F+T & E+R & T & F+T & E+R \\
    \midrule
    TOG~\cite{chow2020adversarial} & .18 / .18 & .31 / \underline{.38} & .17 / .14 & \underline{.20} / .20 & .41 / \underline{.44} & \underline{.24} / .21 & .22 / \underline{.20} & .49 / .34 & \underline{.25} / .15   \\
    TOG+RAND & .19 / .22 & .22 / .29 & .04 / .09 & .18 / .18 & .27 / .35 & .06 / .23 & .22 / \underline{.20} & .32 / .34 & .06 / .15  \\
    TOG+SAME & .18 / .23 & \underline{.45} / .35 & \underline{.21} / \underline{.22} & \underline{.20} / \underline{.22} & \underline{.51} / .43 & .20 / \underline{.27} & .23 / .19 & \underline{.55} / \underline{.38} & .20 / \underline{.26} \\
    Cai~\cite{cai2022context} & \underline{.21} / \underline{.24} & .44 / .30 & .11 / .10 & \underline{.20} / \underline{.22} & .48 / .38 & .11 / .15 & \underline{.24} / .19 & .53 / .37 &.09 / .16 \\
    \midrule
    GLOW & \textbf{.38} / \textbf{.35} & \textbf{.64} / \textbf{.50} & \textbf{.32} / \textbf{.24} & \textbf{.40} / \textbf{.35} & \textbf{.67} / \textbf{.55} & \textbf{.35} / \textbf{.29} & \textbf{.44} / \textbf{.33} & \textbf{.69} / \textbf{.48} & \textbf{.32} / \textbf{.29} \\ \hline\hline
    &\multicolumn{9}{c}{Zero query black-box (coco17val / Pascal)} \\
    \midrule
    TOG~\cite{chow2020adversarial} & .25 / .26 & .04 / .08 & .01 / .04 & .24 / .25 & .08 / .15 & .02 / .12 & .28 / .21 & .12 / .12 & .03 / .15 \\
    TOG+RAND  & .20 / .28 & .05 / .08 & .01 / .03 & .20 / .20 & .08 / .12 & .01 / .05 & .29 / .25 & .14 / .15 & .03 / .07 \\
    TOG+SAME & .23 / .30 & \underline{.12} / \underline{.10} & \underline{.02} / .06 & .22 / .25 & \underline{.20} / \textbf{.19} & \underline{.03} / \textbf{.18} & .27 / \underline{.26} & \underline{.25} / \underline{.17} & \textbf{.05} / .16   \\
    Cai~\cite{cai2022context} & \underline{.26} / \underline{.37} & .10 / .09 & \underline{.02} / \underline{.07} & \underline{.25} / \underline{.26} & .15 / .12 & .02 / .13 & \underline{.30} / .21 & .20 / .14 & .02 / \underline{.18} \\
    \midrule
    GLOW & \textbf{.37} / \textbf{.38} & \textbf{.17} / \textbf{.14} & \textbf{.03} / \textbf{.10} & \textbf{.38} / \textbf{.35} & \textbf{.22} / \underline{.18} & \textbf{.04} / \underline{.15} & \textbf{.44} / \textbf{.38} & \textbf{.29} / \textbf{.23} & \textbf{.05} / \textbf{.19} \\     
    \bottomrule
\end{tabular}

%% file: tables/res_r3_iccv_v1.tex
\begin{tabular}{c|ccc|ccc|ccc}
    \toprule
    \multirow{3}{*}{Methods} 
    &\multicolumn{9}{c}{White-box (coco17val / Pascal)} \\ \cline{2-10}
    &\multicolumn{3}{c|}{R3-5}
    &\multicolumn{3}{c|}{R3-50}
    &\multicolumn{3}{c}{R3-95}\\ \cline{2-10}
    & T & F+T+C & C+R & T & F+T+C & C+R & T & F+T+C & C+R \\
    \midrule
    TOG~\cite{chow2020adversarial} & .18 / .21 & \underline{.35} / \textbf{.28} & .10 / .05 & .19 / .15 & \underline{.43} / \textbf{.37} & \underline{.13} / \textbf{.16} & .22 / .19 & \underline{.50} / \underline{.35} & \textbf{.14} / \textbf{.17} \\
    TOG+RAND & .18 / \textbf{.30} & .27 / .24 & .08 / .05 & .19 / .20 & .33 / .32 & .11 / \textbf{.16} & .22 / .19 & .38 / .31 & .12 / .16   \\
    TOG+SAME & .18 / .23 & .14 / .16 & \underline{.11} / \underline{.06} & .20 / .21 & .15 / .18 & .12 / .14 & .22 / .19 & .17 / .16 & .12 / \textbf{.17} \\
    Cai~\cite{cai2022context} & \underline{.20} / .25 & .17 / .10 & .02 / .02 & \underline{.21} / \underline{.24} & .22 / .16 & .02 / .06 & \underline{.23} / \underline{.22} & .21 / .13 & .02 / .04   \\
    \midrule
    GLOW & \textbf{.32} / \underline{.28} & \textbf{.48} / \underline{.27} & \textbf{.13} / \textbf{.07} & \textbf{.34} / \textbf{.28} & \textbf{.52} / \underline{.35} & \textbf{.15} / .12 & \textbf{.35} / \textbf{.27} & \textbf{.53} / \textbf{.37} & \textbf{.14} / .11  \\ 
    \hline \hline
    &\multicolumn{9}{c}{Zero query black-box (coco17val / Pascal)} \\ 
    \midrule
    TOG~\cite{chow2020adversarial} & .21 / .32 & .01 / .01 & .00 / .01 & .21 / .23 & \underline{.03} / \textbf{.03} & .01 / \underline{.03} & .28 / .26 & \textbf{.04} / \textbf{.04} & \textbf{.02} / \textbf{.05} \\
    TOG+RAND & .22 / .31 & .01 / .01 & .00 / .01 & .21 / .21 & .02 / \textbf{.03} & .01 / \underline{.03} & .28 / .25 & \textbf{.04} / \textbf{.04} & .01 / .04\\
    TOG+SAME & .21 / .32 & .01 / .01 & .00 / .01 & .22 / .23 & .02 / .01 & .01 / \textbf{.04} & .27 / .25 & .03 / .03 & \textbf{.02} / \textbf{.05} \\
    Cai~\cite{cai2022context} & \underline{.25} / \textbf{.35} & .01 / .01 & .00 / .01 & \underline{.25} / \underline{.31} & .01 / .00 & .01 / .01 & \underline{.30} / \underline{.27} & .02 / .02 & .00 / .03  \\
    \midrule
    GLOW & \textbf{.31} / \textbf{.35} & \textbf{.02} / .01 & \textbf{.01} / \textbf{.03} & \textbf{.34} / \textbf{.38} & \textbf{.04} / .02 & .01 / .02 & \textbf{.37} / \textbf{.36} & \textbf{.04} / \textbf{.04} & .01 / .03 \\
    \bottomrule
\end{tabular}

%% file: tex/conclusion.tex
~\vspace{-7mm}
\section{Conclusion}~\label{sec:conclusion}
In this paper, we propose a novel attack generation algorithm GLOW for adversarial attacks on detectors. Compared to existing work, it explicitly takes both semantic context and geometric layout into consideration. By validating on two datasets, we demonstrate that GLOW produces superior performances under both conventional attack request and more generic ones where victim objects are obtained by estimation. 
GLOW also showcases better transfer rates under challenging zero-query black-box setting. 